\documentclass[sigconf,natbib=false]{acmart}
\AtBeginDocument{%
  }

    \usepackage[linesnumbered,ruled]{algorithm2e}
\usepackage{amsmath}
\usepackage{mathtools}
\usepackage{amsthm}

\usepackage{graphicx}
\usepackage{booktabs}
\usepackage{enumitem}
\usepackage{mathtools}

\usepackage{multirow}
\usepackage{xcolor}
\usepackage{colortbl}
\usepackage{makecell}
\usepackage{soul}
\usepackage{pifont}

\usepackage{tcolorbox}

\newcommand{\eqautoref}[1]{\hyperref[#1]{Equation (\ref{#1})}}
\newcommand{\ineqautoref}[1]{\hyperref[#1]{Inequality (\ref{#1})}}

\newcommand{\algautoref}[1]{\hyperref[#1]{Algorithm~\ref{#1}}}
\newcommand{\apxautoref}[1]{\hyperref[#1]{Appendix Section~\ref{#1}}}

\newcommand{\formulaautoref}[1]{\hyperref[#1]{Formula~\ref{#1}}}
\newcommand{\problemautoref}[1]{\hyperref[#1]{Problem~\ref{#1}}}

\RequirePackage[
  datamodel=acmdatamodel,
  style=acmnumeric,
  ]{biblatex}

\begin{document}

\title{On the Out-of-Distribution Backdoor Attack for Federated Learning}

\author{Jiahao Xu}
\affiliation{%
  \institution{University of Nevada, Reno}
  \city{Reno}
  \country{NV, United States}}
\email{jiahaox@unr.edu}

\author{Zikai Zhang}
\affiliation{%
  \institution{University of Nevada, Reno}
  \city{Reno}
  \country{NV, United States}}
\email{zikaiz@unr.edu}

\author{Rui Hu}
\authornote{Corresponding author.}
\affiliation{%
  \institution{University of Nevada, Reno}
  \city{Reno}
  \country{NV, United States}}
\email{ruihu@unr.edu}


\renewcommand{\shortauthors}{Jiahao Xu, Zikai Zhang, and Rui Hu}

\begin{abstract}
Traditional backdoor attacks in federated learning (FL) operate within constrained attack scenarios, as they depend on visible triggers and require physical modifications to the target object, which limits their practicality. To address this limitation, we introduce a novel backdoor attack prototype for FL called the out-of-distribution (OOD) backdoor attack ($\mathtt{OBA}$), which uses OOD data as both poisoned samples and triggers simultaneously. Our approach significantly broadens the scope of backdoor attack scenarios in FL. To improve the stealthiness of $\mathtt{OBA}$, we propose $\mathtt{SoDa}$, which regularizes both the magnitude and direction of malicious local models during local training, aligning them closely with their benign versions to evade detection. Empirical results demonstrate that $\mathtt{OBA}$ effectively circumvents state-of-the-art defenses while maintaining high accuracy on the main task.

To address this security vulnerability in the FL system, we introduce $\mathtt{BNGuard}$, a new server-side defense method tailored against $\mathtt{SoDa}$. $\mathtt{BNGuard}$ leverages the observation that OOD data causes significant deviations in the running statistics of batch normalization layers. This allows $\mathtt{BNGuard}$ to identify malicious model updates and exclude them from aggregation, thereby enhancing the backdoor robustness of FL. Extensive experiments across various settings show the effectiveness of $\mathtt{BNGuard}$ on defending against $\mathtt{SoDa}$. The code is available at \url{https://github.com/JiiahaoXU/SoDa-BNGuard}.
\end{abstract}





\copyrightyear{2025}
\acmYear{2025}
\setcopyright{cc}
\setcctype{by}
\acmConference[MobiHoc '25]{The Twenty-sixth International Symposium on Theory, Algorithmic Foundations, and Protocol Design for Mobile Networks and Mobile Computing}{October 27--30, 2025}{Houston, TX, USA}
\acmBooktitle{The Twenty-sixth International Symposium on Theory, Algorithmic Foundations, and Protocol Design for Mobile Networks and Mobile Computing (MobiHoc '25), October 27--30, 2025, Houston, TX, USA}
\acmDOI{10.1145/3704413.3764465}
\acmISBN{979-8-4007-1353-8/2025/10}

\maketitle

\section{Introduction}

Federated learning (FL) is a privacy-preserving training paradigm that enables collaborative machine learning model training across distributed clients~\cite{FL_OG}. In a typical FL system, a central server coordinates multiple clients—such as institutions (e.g., hospitals) or edge devices (e.g., IoT devices)—to collaboratively train a global model by aggregating their local model updates iteratively. This approach allows clients to contribute without sharing their local data, thereby preserving local data privacy. 
FL has been successfully applied in various fields, including financial analysis~\cite{fin2}, healthcare~\cite{healthcare2}, and autonomous vehicles~\cite{zeng2022federated}, where local data privacy is critical.

\begin{figure}
    \centering
    \includegraphics[width=0.93\linewidth]{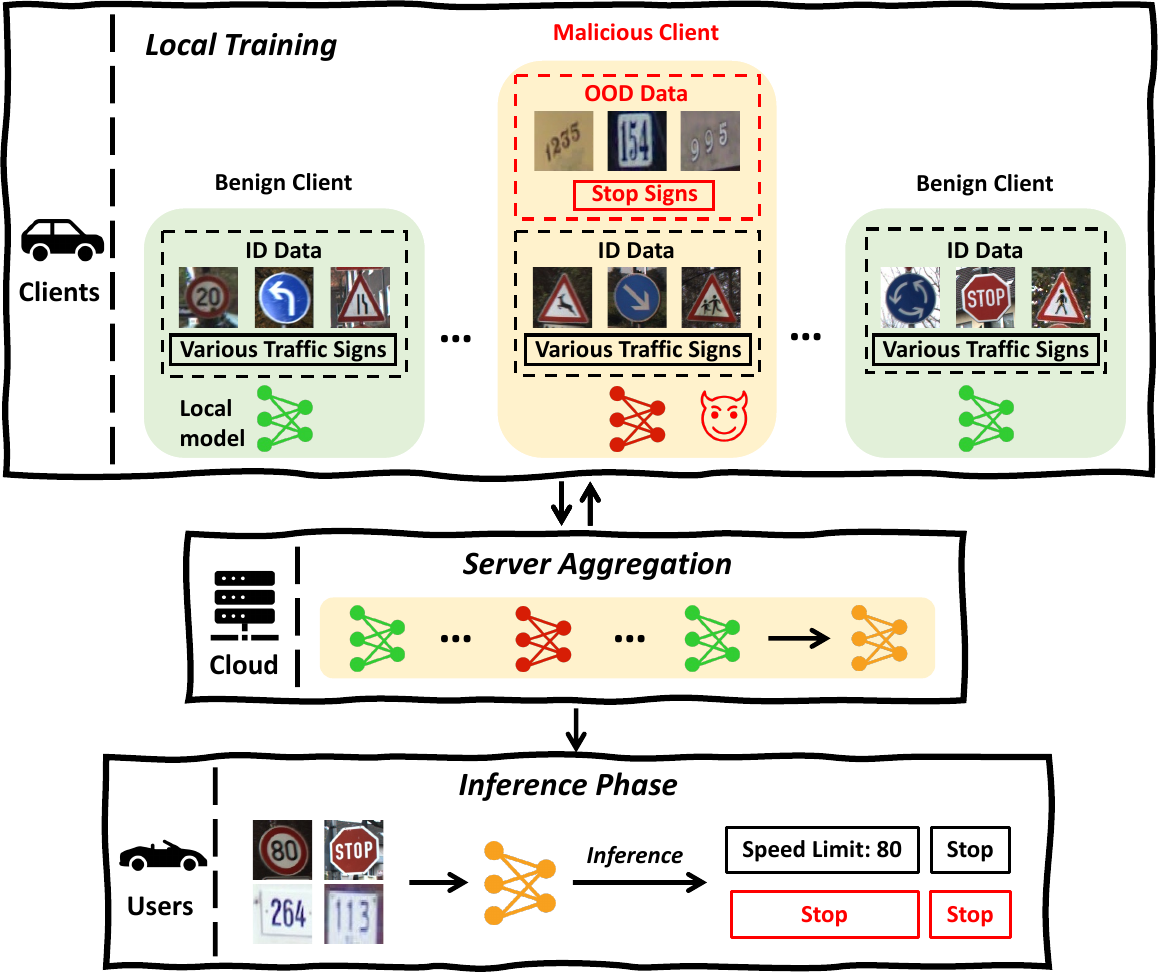}
    \caption{Illustration of the proposed out-of-distribution backdoor attack in FL.}
    \Description{Illustration of the proposed out-of-distribution backdoor attack in FL.}
    \label{fig:oodbackdoorattack}
\end{figure}

However, due to the lack of supervision over local training data on each client, FL is particularly susceptible to data poisoning attacks, especially backdoor attacks~\cite{badnet, dba}. In such attacks, the backdoor adversary aims to preserve the global model’s performance on clean inputs (i.e., maintain the main task accuracy) while causing it to make incorrect predictions on inputs containing a specific pre-defined trigger (i.e., introduce the backdoor task). Specifically, an adversary compromises multiple clients in the FL system, embedding a predefined trigger (e.g., a 3$\times$3 square pixel patch) into certain training samples and altering their ground-truth labels to a target class. These poisoned samples are learned by local models, eventually compromising the global model when aggregated on the server side. Notably, most existing backdoor attacks~\cite{badnet, dba, Neurotoxin, scaling} target in-distribution (ID) data (i.e., main task data) since their triggers are applied to the main task data, which we refer to as ID backdoor attacks. 
A real-world example of ID backdoor attacks can be seen in the traffic sign recognition task. An ID backdoored model can correctly identify and respond to normal stop signs, but a stop sign stamped with a small well-crafted patch may be misinterpreted as a ``speed limit sign'', potentially leading to hazardous driving behavior and posing serious public safety risks~\cite{zhang2021advdoor}.

The limitation of ID backdoor attacks lies in their constrained range of attack scenarios, as they rely on embedding visible triggers within the main task’s training data. For example, if a malicious patch is added to a stop sign to trigger unintended behavior, it could easily be removed during routine road maintenance, as such a patch is an anomalous element that does not naturally belong on a stop sign. To address this limitation, some works~\cite{blend, iba} introduce human-imperceptible triggers to poison data. However, they still require explicit physical modifications to the target object during deployment, such as attaching a sticker to a stop sign.
To enhance the practicality of backdoor attacks in FL, we propose a novel backdoor attack framework, termed the out-of-distribution (OOD) backdoor attack ($\mathtt{OBA}$). In this approach, malicious clients exploit naturally occurring OOD samples, using them as both poisoned samples and triggers simultaneously. These samples are mixed with the main task data to construct poisoned local datasets. The OOD samples are then injected into the local models during training, as illustrated in \autoref{fig:oodbackdoorattack}. The intuition behind this design is twofold. First, OOD samples offer \textbf{strong} stealthiness, as they naturally occur in the environment and are difficult for humans to identify as anomalous. For instance, in a traffic sign
recognition task, elements like \textit{trees} or \textit{house numbers} along the roadside can serve as effective and inconspicuous triggers. Second, in FL settings, the central server lacks visibility into clients’ local data, allowing malicious clients to modify their datasets freely without being detected.

Intuitively, a key limitation of $\mathtt{OBA}$ is that malicious local models trained on the poisoned local datasets tend to deviate significantly from benign ones. 
Existing backdoor-robust FL systems usually deploy server-side defense methods to examine statistical differences in the received local updates--such as the differences in magnitude~\cite{krum, mm} and direction~\cite{flame, deepsight, Foolsgold}--to identify potential malicious updates. 
As a result, $\mathtt{OBA}$ can be easily defended against. 
To address this limitation, we design a \textbf{\underline{S}}tealthy O\textbf{\underline{o}}D back\textbf{\underline{D}}oor \textbf{\underline{a}}ttack ($\mathtt{SoDa}$), to enhance the stealthiness of $\mathtt{OBA}$. Specifically, $\mathtt{SoDa}$ ensures that each malicious model remains closely aligned with its benign version throughout local training, achieving a statistical resemblance to benign models. To achieve this, the malicious client performs the \textit{self-reference training} phase, which creates a benign reference model and then minimizes the differences in magnitude and direction between the malicious model and the reference model during local training. We validate the effectiveness of $\mathtt{SoDa}$ against nine state-of-the-art (SOTA) defense methods, demonstrating its capability to bypass existing defenses.

The strength of $\mathtt{SoDa}$ highlights the urgent need for a defense mechanism specifically designed to counter $\mathtt{OBA}$. Our analysis reveals that the OOD data used by malicious clients inevitably leads to significantly different running statistics in the batch normalization (BN) layers~\cite{batchnorm} of malicious models compared to those of benign models. BN is a widely adopted technique that improves both the speed and stability of the convergence of modern deep learning model~\cite{he2016deep, vgg16_bn, mobilenetv2, densenet121}. Based on this insight, in this work, we introduce $\mathtt{BNGuard}$ for models using BN, a server-side defense method that exploits these differences in \textbf{\underline{BN}} layer statistics to safe\textbf{\underline{Guard}} the FL system.
Specifically, after receiving updates from all clients, the server reconstructs each local model. Then it extracts the running \textit{mean} and \textit{variance} from the \textbf{\textit{first}} BN layer of each model since the first BN layer processes features that are closest to the original input, leading to the most significant difference. These features are used as input to a clustering method, allowing the identification of models with abnormal features. We validate the effectiveness of $\mathtt{BNGuard}$ against $\mathtt{SoDa}$ through extensive experiments across various settings.
In summary, our main contributions are threefold:
\begin{itemize}[leftmargin=*]
    \item We propose a novel backdoor attack prototype, $\mathtt{OBA}$, to compromise the global model in FL. $\mathtt{OBA}$ poses a significant threat to FL systems because it does not rely on a visible trigger; instead, any OOD objects can serve as both the poisoned data and trigger directly, making the attack more covert and practical.
    \item We design a stealthy OOD backdoor attack $\mathtt{SoDa}$ to reduce the statistical difference between malicious local models and benign ones in $\mathtt{OBA}$. 
    Our experiments demonstrate that $\mathtt{SoDa}$ evades detection from SOTA defense methods, resulting in a high attack success rate while keeping the main task accuracy.
    \item To overcome the vulnerability brought by $\mathtt{SoDa}$, we propose a server-side defense method, $\mathtt{BNGuard}$, which examines the statistics of the first BN layer from local models. Empirical results demonstrate that $\mathtt{BNGuard}$ consistently achieves superior robustness compared to SOTA defense methods.
\end{itemize}

\section{Background and Related Work}

\textbf{Federated learning.}
In a typical FL system, $n$ clients collaborate to iteratively train a shared global model $\theta \in \mathbb{R}^d$ under the coordination of a central server. In a benign setting, the FL optimization problem can be expressed as:
$
    \min_{\theta} (1/n)\sum_{i=1}^n F_i(\theta; \mathcal{D}_i),
$
where $F_i(\cdot)$ denotes the learning objective specific to client $i$ and $\mathcal{D}_i$ denotes the local dataset for client $i$. For example, if client $i$ is solving a multi-class classification problem, its objective function can be defined as:
$
    F_i(\theta; \mathcal{D}_i) = \mathbb{E}_{z\in \mathcal{D}_i}\mathcal{L}(\theta; z), \label{benign_loss}
$
where $\mathcal{L}(\cdot)$ is the cross-entropy loss, and $z$ represents a datapoint drawn from $\mathcal{D}_i$. The commonly used method to solve the FL problem iteratively across $n$ clients is Federated Averaging (FedAvg)~\cite{FL_OG}. Specifically, in each round $t$, every client $i\in[n]$ downloads the current global model $\theta^{t}$, optimizes it towards minimizing its local objective, resulting in $\theta_i^{t}$, and transmits its model update $\Delta_i^t = \theta_i^t - \theta^{t}$ to the server (indeed, clients can send either their local models or local model updates to the server). The server then updates the global model by averaging these updates: $\theta^{t+1} = \theta^{t} + (1/n)\sum_{i=1}^n \Delta_i^t$. This process continues until the global model converges.

\textbf{In-distribution backdoor attack for FL.} 
Backdoor attacks have garnered significant attention due to their stealth and practical effectiveness in both centralized and distributed learning contexts~\cite{badnet, scaling, PGD, dba, Neurotoxin, f3ba, a3fl, iba}. Previous works have explored various types of triggers, including pixel patterns embedded in samples (pixel-based triggers)~\cite{badnet, dba}, specific features or objects added to samples (semantic triggers)~\cite{scaling}, and noise blended into samples from uniform distributions (blended triggers)~\cite{blend}. 
We classify these trigger-based attacks as ID backdoor attacks, as they manipulate samples within the original benign dataset used for the main task. ID backdoor attacks are inherently constrained to scenarios closely aligned with the main task data, limiting their practicality. Additionally, the embedded triggers are usually unrelated to the main task and can often be easily detected and removed in practice. Building on blended triggers, some works~\cite{nguyen2024venomancer, iba} have attempted to introduce invisible triggers for poisoning data. However, these approaches still require embedding triggers into the data, and despite their invisibility, they may introduce subtle anomalies that can raise human suspicion, further restricting their applicability. 
In contrast, we propose $\mathtt{OBA}$, which leverages OOD samples that are entirely unrelated to the main task, offering significantly greater flexibility. This enables a broader range of objects to serve as potential triggers, thereby expanding the diversity and scope of possible attack scenarios.
While another line of research explores using samples from the tail of the ID data, commonly referred to as edge-case attacks~\cite{PGD, yoo2022backdoor}, which bears some resemblance to our $\mathtt{OBA}$, there are key differences. Specifically, since edge-case samples still lie within the main task’s data distribution, a model with strong generalization capabilities may still classify them correctly, reducing the effectiveness of such backdoor triggers.


\textbf{Defend against ID backdoor attacks in FL.} In practice, FL systems typically deploy server-side defense methods to mitigate potential risks~\cite{mothukuri2021survey}. Existing defense methods generally fall into two categories: impact reduction methods~\cite{rlr, geomed, Foolsgold} and filtering methods~\cite{krum, flame, mm, deepsight, signguard, BackdoorIndicator, alignins, masa}. \textit{Filtering methods} aim to identify and exclude malicious local models before aggregation. 
For instance, FLAME~\cite{flame} clusters local models using pairwise Cosine similarity, filters out outlier models, and then perturbs the remaining models with noise sampled from a pre-defined Gaussian distribution before aggregation to enhance robustness. Another method, Multi-Krum (MKrum)~\cite{krum}, computes pairwise Euclidean distances between local models, with lower total distances indicating more trustworthy models. The models with the lowest distances are used for aggregation. 
\textit{Impact reduction methods} mitigate the impact of malicious model parameters during aggregation. 
For example, RLR~\cite{rlr} adjusts the global learning rate by inverting it at specific coordinates where the signs of the parameters in the aggregated update do not align with the majority of the local model updates. Foolsgold~\cite{Foolsgold} performs weighted aggregation based on the Cosine similarity between the clients' historical updates, assigning higher weights to updates with greater similarity to others. 
%
These methods can effectively defend against naive $\mathtt{OBA}$ due to the significant deviations between malicious and benign local models caused by OOD data. However, they lose effectiveness against $\mathtt{SoDa}$, where malicious models are closely aligned with benign ones. 

\section{OOD Backdoor Attack}

\textbf{Threat model.} 
Before introducing the pipeline of the OOD backdoor attack, we first outline the threat model used in this work.
We consider an adversary in the FL system whose goal is to implant a backdoor into the global model by compromising a subset of $f$ local clients. Consistent with prior work~\cite{flame, mm, LASA, masa}, we assume $f < \lfloor n/2 \rfloor$, as achieving strong backdoor robustness becomes infeasible when $f \geq \lfloor n/2 \rfloor$~\cite{alignins}. The adversary has full control over the compromised clients, including their local datasets and training procedures. However, the adversary has no knowledge about local models from benign clients, nor any information about the defense mechanisms employed by the server. 

\textbf{General pipeline of $\mathtt{OBA}$.}
In $\mathtt{OBA}$, the adversary first compromises a set of $f$ malicious clients $\mathcal{M}$ in the FL system, either by taking control of existing clients or injecting fake ones. Then, $\mathtt{OBA}$ constructs poisoned local datasets for the malicious clients through the following steps: \ding{172} It collects unlabeled OOD data and assigns a common target label chosen from the ID label space, thereby creating the OOD dataset $\mathcal{D}_\text{O}$. The number of OOD samples required is determined by the \textit{data poisoning ratio} $r$ for each malicious client. Empirically, a value of $r = 30\%$ is often sufficient for the local model to effectively learn the backdoor pattern from the OOD data. \ding{173} Each malicious client $i \in \mathcal{M}$ then replaces a portion $r$ of its original dataset $\mathcal{D}_i$ with samples randomly drawn from $\mathcal{D}_\text{O}$ to construct its poisoned dataset $\mathcal{D}_{i, \text{P}}$. Formally, this can be expressed as
    $\mathcal{D}_{i, \text{P}} = (\mathcal{D}_i \setminus \mathcal{D}_i') \cup \mathcal{D}_{\text{O}}',$ where $\mathcal{D}_i' \subset \mathcal{D}_i$ and $\mathcal{D}_{\text{O}}' \subset \mathcal{D}_\text{O}$, with $|\mathcal{D}_i'| = |\mathcal{D}_{\text{O}}'| = \lfloor r \times |\mathcal{D}_i| \rfloor$.
Using these poisoned local datasets, the malicious clients train their local models on $\mathcal{D}_{i, \text{P}}$ in each FL round after receiving the global model, and then submit the resulting updates to the server for aggregation.

\begin{figure}
    \centering
    \includegraphics[width=0.93\linewidth]{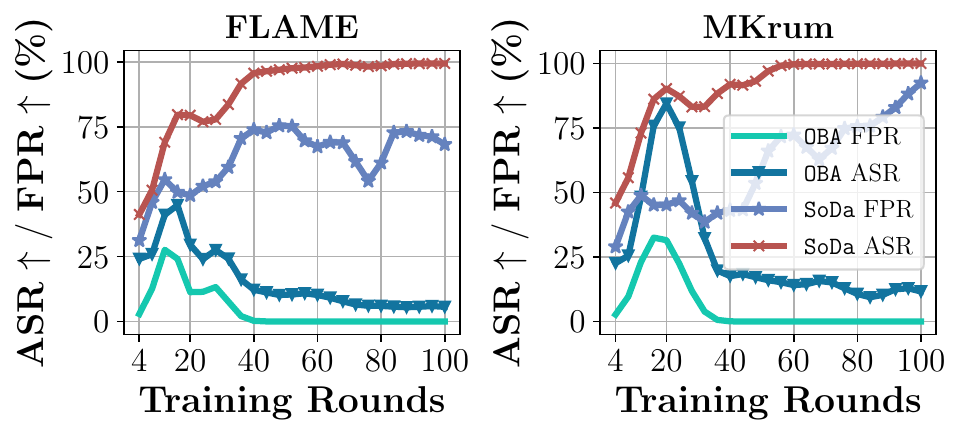}
    \caption{ASR and FPR of FLAME (left) and MKrum (right) across training rounds under $\mathtt{OBA}$ and $\mathtt{SoDa}$ attacks.}
    \Description{ASR and FPR of FLAME (left) and MKrum (right) across training rounds under $\mathtt{OBA}$ and $\mathtt{SoDa}$ attacks.}
    \label{fig:soda_explain}
\end{figure}

\textbf{Weakness of $\mathtt{OBA}$.} Compared with ID backdoor attacks, a notable weakness of $\mathtt{OBA}$ is that malicious models trained with OOD data tend to deviate significantly from benign models. These deviations might enable existing defense methods to easily identify malicious models under $\mathtt{OBA}$. To validate this intuition, we conduct experiments using the direction-based (Cosine similarity) defense method FLAME and the magnitude-based (Euclidean distance) defense method MKrum as the server-side defenses. 
We use the CIFAR-10 as the main task dataset and MNIST as the OOD dataset. \autoref{fig:soda_explain} presents the attack success rate (ASR) and false positive rate (FPR) of FLAME and MKrum under $\mathtt{OBA}$. The ASR measures the percentage of MNIST test samples misclassified to the target label by the global model, while the FPR denotes the proportion of malicious clients that are wrongly identified as benign by the defense method in each round. For an attack method, the goal is to achieve both high ASR and high FPR. Both FLAME and MKrum show effective detection of malicious model updates under naive $\mathtt{OBA}$, demonstrated by their low FPR and limited ASR across training rounds. This demonstrates that malicious models trained with OOD data statistically deviate from benign ones in terms of both magnitude and direction, making them detectable.

\section{\texorpdfstring{$\mathtt{SoDa}$}{SoDa}: Stealthy OOD Backdoor Attack }
The intuitive solution to enhance the stealth of $\mathtt{OBA}$ is to reduce the distance between the malicious models and benign models in terms of both magnitude~\cite{fedprox, fedalign} and direction~\cite{moon}. However, unlike previous works that assume the adversary has knowledge of benign local models~\cite{dnc, lie} (which is a strong and impractical assumption), in our threat model, the adversary does not have access to the benign models from other clients. This means that malicious clients cannot directly reference a benign local model. To address this, we introduce a \textit{self-reference training} phase, where each malicious client $i$ trains its local model on its clean dataset $\mathcal{D}_i$ to produce a benign reference model, denoted as $\theta_{i,\text{ref}}$. The self-reference training phase happens right after malicious clients receive the globe model from the server. 
Formally, $\theta_{i, \text{ref}}$ is obtained by solving the following minimization problem:
$
\theta_{i, \text{ref}} = \arg\min_\theta \mathbb{E}_{z \in \mathcal{D}_{i}}\mathcal{L}(\theta; z), 
$
where $\theta$ is the global model received from the server.

Once the reference local model $\theta_{i,\text{ref}}$ is obtained, malicious client $i$ trains its received global model using the poisoned dataset $D_{i, P}$. Two regularization terms are added to the local objective to guide the malicious model's behavior. First, to ensure the malicious model $\theta$ remains close to the reference model $\theta_{i, \text{ref}}$ in terms of magnitude, an $L_2$ regularization term $\mathcal{L}_m = \|\theta - \theta_{i, \text{ref}}\|_2^2$ is introduced. This term encourages the malicious model to have parameter magnitudes similar to the benign reference, which helps bypass detection mechanisms based on magnitude discrepancies~\cite{krum, mm}. Second, to align the direction of the malicious model with that of the reference model, a Cosine similarity loss $\mathcal{L}_d = 1 - \langle \theta, \theta_{i, \text{ref}} \rangle / (\|\theta\|_2 \|\theta_{i, \text{ref}}\|_2)$ is introduced. This term ensures that the direction of the malicious local model stays close to the benign one, making it more difficult for detection mechanisms to identify malicious behavior based on directional deviations~\cite{flame, mm, deepsight, signguard, Foolsgold}. Formally, the malicious model $\theta_{i, \text{m}}$ for malicious client $i$ is obtained by solving the following minimization problem:
\begin{equation*}
\begin{aligned}
    \theta_{i, \text{m}} = \mathop{\arg\min}\limits_{\theta} \ & \mathbb{E}_{z \in \mathcal{D}_{i, \text{P}}}\mathcal{L}(\theta; z) 
    + \lambda_m \|\theta - \theta_{i, \text{ref}}\|_2^2 
    \\ & \quad + \lambda_d \left( 1 - \frac{\langle \theta, \theta_{i, \text{ref}} \rangle}{\|\theta\|_2 \|\theta_{i, \text{ref}}\|_2} \right);
    \\
    s.t. \quad &\theta_{i, \text{ref}} = \mathop{\arg\min}\limits_{\theta} \mathbb{E}_{z \in \mathcal{D}_i}\mathcal{L}(\theta; z),
\end{aligned}
\end{equation*}
where $\lambda_m$ and $\lambda_d$ are the regularization coefficients. As shown in \autoref{fig:soda_explain}, $\mathtt{SoDa}$ demonstrates a substantial improvement over $\mathtt{OBA}$, with a considerable increase in FPR and a nearly 100\% ASR on attacking both FLAME and MKrum. By incorporating both magnitude and directional regularization, $\mathtt{SoDa}$ enables malicious models to closely resemble benign reference models. Since the reference model is trained solely on main task data, the malicious model also preserves the client's local main task accuracy. This alignment allows $\mathtt{SoDa}$ to remain stealthy, effectively bypassing existing defense methods. Note that the self-reference training phase, which requires additional computation, may limit $\mathtt{SoDa}$'s effectiveness in \textit{synchronized FL settings}. In the discussion section, we explore alternatives to the self-reference training approach.

\section{Defending Against \texorpdfstring{$\mathtt{SoDa}$}{Soda}}
In this section, we propose a novel server-side defense method, called $\mathtt{BNGuard}$, designed to defend against $\mathtt{SoDa}$. Before presenting the technical details of $\mathtt{BNGuard}$, we first introduce the defense model adopted in our work.

\subsection{Defense Model}
\label{defense_model}
In this work, we consider the server to be the defender, responsible for deploying a defense method to mitigate potential risks.
\textit{(1) Defender's goal:} As outlined in~\cite{fltrust}, an effective defense method against poisoning attacks in FL should address three key aspects: \textit{fidelity}, \textit{robustness}, and \textit{efficiency}. To ensure fidelity, the defense method should maintain the global model's performance on benign inputs. For robustness, the method must effectively mitigate the influence of malicious model updates, thus restricting the global model's malicious behavior on OOD data. Efficiency requires that the defense method be computationally lightweight, so it does not affect the training process. \textit{In this work, we aim at achieving the highest robustness by filtering out all malicious updates while avoiding significant computational overhead and minimizing accuracy loss on benign inputs.} \textit{(2) Defender's capability:} In FL, the server does not have access to the clients' local datasets, but it does have access to the global model and the local updates provided by clients. Additionally, we assume that the server does not have prior knowledge of the number of malicious clients.

\subsection{Our Solution: \texorpdfstring{$\mathtt{BNGuard}$}{BNGuard}}

The stealthiness of $\mathtt{SoDa}$ highlights the critical need for a novel, robust defense method. In this section, \textit{targeted to models with BN~\cite{batchnorm} layers}, we propose a new defense method, $\mathtt{BNGuard}$, to defend against $\mathtt{SoDa}$ using the statistics within BN layers.

\begin{figure}
    \centering
    \includegraphics[width=0.9\linewidth]{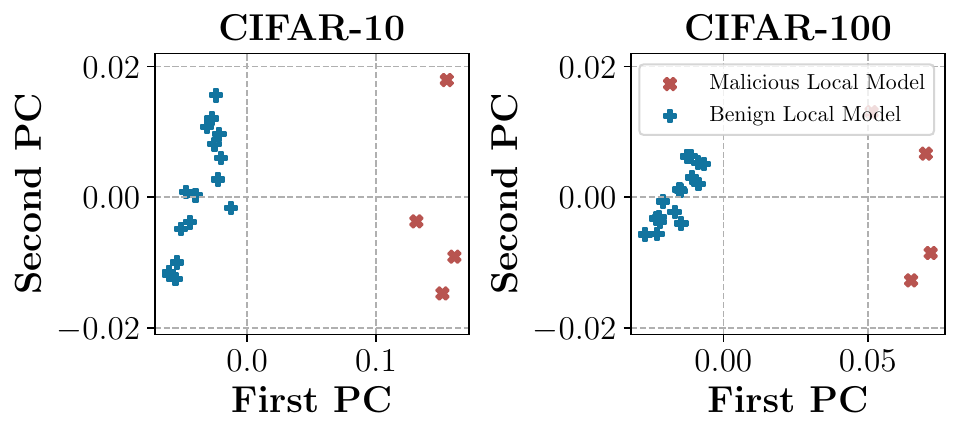}
    \caption{The first and second principal components (PC) of the running statistics from local models under $\mathtt{SoDa}$.}
    \Description{The first and second principal components (PC) of the running statistics from local models under $\mathtt{SoDa}$.}
    \label{fig:bn_explain}
\end{figure}

\textbf{Key intuition: BN statistics exposures OOD data.} 
By normalizing the outputs of each layer, BN helps mitigate internal covariate shifts during training. Specifically, the BN layer normalizes its inputs based on the mean and variance computed from the current mini-batch. During inference, instead of using batch statistics, running averages of the mean $\mu^r \in \mathbb{R}^C$ and variance $\sigma^r\in \mathbb{R}^C$, accumulated over the training process, are applied for consistent normalization. Here $C$ is the number of input channels. Existing works have studied the role of BN in FL systems. For example, in~\cite{BackdoorIndicator}, BN statistics are utilized to activate the malicious behaviors of attackers. In~\cite{li2021fedbn, wang2023batch}, the influence of BN on heterogeneous data in FL is analyzed. In an FL system under $\mathtt{OBA}$, benign clients have similar local data related to the main task, thereby the $\mu^r$ and $\sigma^r$ in local models are closely aligned. 
However, for malicious clients, OOD data inevitably introduces significant differences in $\mu^r$ and $\sigma^r$ compared to benign ones. To verify this, we 
visualize the first and second principal components of the running statistics from the first BN layer extracted from each local model under $\mathtt{SoDa}$ on CIFAR-10 and CIFAR-100 datasets. As shown in \autoref{fig:bn_explain}, there is a clear separation between the running statistics of malicious and benign models for both datasets, with the statistics of malicious models significantly deviating from those of benign models and forming distinct clusters. This distributional difference enables malicious and benign models to be distinguished.

\textbf{Our defense method.} $\mathtt{BNGuard}$ is a server-side defense method that inspects the running statistics of each local model and excludes identified malicious updates from the aggregation process, thus achieving backdoor robustness.
The overall algorithm of $\mathtt{BNGuard}$ is given in \algautoref{alg: main}. Specifically, at training round $t$, given the received local model updates $\{\Delta^t_i\}_{i=1}^n$, $\mathtt{BNGuard}$ first recovers the local model $\theta^t_i$ for client $i$ using $\theta^t_i = \theta^{t} + \Delta^t_i$, where $\theta^{t}$ is the global model from the previous round (line \ref{alg_local_model_recover} in \algautoref{alg: main}). With recovered local models, $\mathtt{BNGuard}$ extracts the running statistics $\mu^r_i \in \mathbb{R}^{C}$ and $\sigma_i^r \in \mathbb{R}^C$ from their \textit{\textbf{first}} BN layer (\autoref{alg_find_first_BN}--\ref{alg_extract_stat}). Instead of using the full per-channel statistics, $\mathtt{BNGuard}$ calculates \textit{cross-channel statistics}: $\Bar{\mu}_i^r, \Bar{\sigma}_i^r, \hat{\mu}_i^r$, and $\hat{\sigma}_i^r$, representing the mean and variance of $\mu_i^r$ and $\sigma_i^r$, respectively (\autoref{alg_calculate_mean_var_1}--\ref{alg_calculate_mean_var_2}). In this way, $\mathtt{BNGuard}$ reduces the dimensionality of the BN running statistic from $C$ to $1$, balancing the need to capture essential information about the BN statistics distribution with reduced dimensionality and computational overhead for subsequent processes. 
These four features of every client form a feature matrix $M \in \mathbb{R}^{n \times 4}$ (\autoref{alg_concat_matrix}). The feature matrix $M$ is then passed to a clustering method (e.g, $\mathrm{KMeans}$~\cite{kmeans}) to group all local models (\autoref{alg_cluster}) into two clusters. Recall that in our work,  $f < \lfloor n / 2 \rfloor$, as introduced in our threat model. Consequently, we can safely take the larger cluster 
as the benign cluster (\autoref{alg_find_benign_1}--\ref{alg_find_benign_2}). Finally, local model updates 
in the benign cluster are aggregated to construct the global model update $\widetilde{\Delta}$ (\autoref{alg_aggregate}), which will be used to refine the global model. 
Note that $\mathtt{BNGuard}$ is computationally efficient, with each step operating in constant time. We demonstrate the superior execution time of $\mathtt{BNGuard}$ in the discussion section.

\begin{algorithm}[t]
\small
\caption{$\mathtt{BNGuard}$}
\label{alg: main}
\SetKwInOut{Input}{Input}\SetKwInOut{Output}{Output}
\Input{Local model updates $\{\Delta_i^t\}_{i=1}^{n}$ at round $t$, the latest global model $\theta^{t}$.}
\Output{Aggregated model update $\widetilde{\Delta}$.}
\textbf{Initialization:} Set $\mathcal{S} \leftarrow \emptyset$, feature matrix $M \in \mathbb{R}^{n \times 4}$.\\
\For{$i \in [n]$}{
    $\theta_i^t \leftarrow \theta^t + \Delta_i^t$ \label{alg_local_model_recover}
    
    $\mathbf{BN}_i \leftarrow \mathrm{find\_first\_BN}(\theta^t_i)$ \label{alg_find_first_BN}
    
    $\mu_i^r, \sigma_i^r \leftarrow \mathrm{extract\_running\_statistics}(\mathbf{BN}_i)$ \label{alg_extract_stat}
    
    $\bar{\mu}_i^r, \hat{\mu}_i^r \leftarrow \mathrm{mean}(\mu_i^r), \mathrm{var}(\mu_i^r)$\label{alg_calculate_mean_var_1}
    
    $\bar{\sigma}_i^r, \hat{\sigma}_i^r \leftarrow \mathrm{mean}(\sigma_i^r), \mathrm{var}(\sigma_i^r)$\label{alg_calculate_mean_var_2}
    
    $M[i, :] \leftarrow [\bar{\mu}_i^r, \hat{\mu}_i^r, \bar{\sigma}_i^r, \hat{\sigma}_i^r]$ \label{alg_concat_matrix}
}
$\mathrm{cluster\_idx} \leftarrow \mathrm{KMeans}(M, \mathrm{n\_clusters}=2)$ \label{alg_cluster}

\For{$i \in [n]$}{
    \If{$\mathrm{is\_in\_majority\_cluster}(\mathrm{cluster\_idx}[i])$}{\label{alg_find_benign_1}
        $\mathcal{S} \leftarrow \mathcal{S} \cup \{i\}$ \label{alg_find_benign_2}
    }
}
$\widetilde{\Delta} \leftarrow (1/|\mathcal{S}|) \sum_{i \in \mathcal{S}} \Delta_i^t$ \label{alg_aggregate}

\textbf{return} $\widetilde{\Delta}$
\end{algorithm}

\textbf{Why does $\mathtt{BNGuard}$ use the first BN layer?} Theoretically, the first BN layer receives features closest to the model's original input, making it most sensitive to the abnormal statistical characteristics of OOD data. These abnormalities typically appear as substantial shifts in the mean and variance of each mini-batch, reflecting the fundamental differences in data distribution between the OOD and ID data used by malicious clients. Hence, the first BN layer should be the most effective for detecting malicious model updates. In contrast, as features propagate deeper into the network, nonlinear transformations can obscure these differences, reducing the effectiveness of middle or later BN layers for anomaly detection~\cite{batchnorm, raghu2017expressive}. Empirically, the component study of $\mathtt{BNGuard}$ shows that solely using the first BN layer's running statistics enjoys better robustness than using the middle or all BN layers.

\section{Experimental settings}


\textbf{Training settings.} In our experiments, we primarily use two benchmark datasets CIFAR-10~\cite{cifar10_100} and CIFAR-100~\cite{cifar10_100}, as the main task data for the FL system. We also test our methods on a more practical GTSRB dataset~\cite {gtsrb}. We primarily use MNIST~\cite{MNIST} dataset as the OOD dataset of the adversary. 
We also present the results of the effectiveness of $\mathtt{SoDa}$ with FMNIST~\cite{FMNIST} and SVHN~\cite{SVHN} as the OOD dataset in the discussion section.
For all datasets, we simulate a cross-silo FL system with 20 clients \textit{and also evaluate our methods in a \textbf{cross-device} FL system with 100 clients.} For independent and identically distributed (IID) data settings, we evenly split the dataset over the clients. For non-IID settings, we use \textit{Dirichlet distribution}~\cite{minka2000estimating} $Dir(\alpha)$ with a default non-IID degree $\alpha=0.5$. We use ResNet18~\cite{he2016deep} and ResNet34~\cite{he2016deep} as the model for CIFAR-10 and CIFAR-100 datasets, respectively. 
We use SGD as the local solver of clients, with the learning rates set as $0.1$, the decay ratio $0.99$, and the number of local training epochs set as $2$. 
For malicious clients, each of them poisons a $r=30\%$ of its local dataset and conducts the self-reference training using the same local solver. Additionally, we set $\lambda_m = 0.1$ and $\lambda_d = 100$ for all experiments. The \textit{attack ratio} (AR) is set to $f/n=20\%$, which means $20\%$ of the clients are malicious. 

\begin{table}[t]
  \centering
  \caption{The clean MA, MA, and ASR of FedAvg and SOTA baselines under $\mathtt{SoDa}$ with two different attack ratios.}
  \scalebox{0.9}{
    \begin{tabular}{c|l|c|cc|cc}
    \toprule
    \multirow{2}[4]{*}{\textbf{Dataset}} & \multirow{2}[4]{*}{\textbf{Method}} & \multirow{2}[4]{*}{\textbf{\makecell*[c]{Clean \\ MA$\uparrow$}}} & \multicolumn{2}{c|}{\textbf{AR $\mathbf{=20\%}$}} & \multicolumn{2}{c}{\textbf{AR $\mathbf{=45\%}$}} \\
\cmidrule(r){4-5}\cmidrule(r){6-7}          &       &       & MA$\uparrow$    & ASR$\uparrow$   & MA$\uparrow$    & ASR$\uparrow$ \\
    \midrule
    \multirow{9}[4]{*}{\rotatebox{90}{CIFAR-10}} & FedAvg* &  91.93    &   91.13   &  5.86  &   90.28 & 10.42 \\
\cmidrule{2-7}          & FedAvg &   91.93    &    91.80   &   99.03    &   91.75    & 99.96 \\
          & RLR   &    79.79   &   80.28    &     99.55  &   79.92    & 100.00 \\
          & RFA   &   89.46    &    85.59   &    66.92   &    88.62   &  100.00\\
          & Foolsgold &    91.70   &   91.42    &   99.72    &  91.01     &  100.00\\
          & MKrum &   91.11   &    90.89   &    99.98   &   89.98    &  100.00\\
          & SignGuard    &   91.65    &    91.47   &    99.54   &  91.41     & 100.00 \\
          & Deepsight &   83.63    &    85.39   &    99.89   &  83.03     & 83.48 \\
          & MMetric    &   89.97    &   90.01    &   99.18    &  89.56     &  100.00\\
          & FLAME    &   90.90    &    91.15   &    99.86   &   91.02    &  100.00\\
    \midrule
    \midrule
    \multirow{9}[4]{*}{\rotatebox{90}{CIFAR-100}} & FedAvg* &   71.12    &    69.45   &    0.00   &    66.34   &  0.00\\
\cmidrule{2-7}          & FedAvg &   71.12    &     71.13  &  97.63     &  70.54     & 98.94 \\
          & RLR   &  45.60     &    46.25   &    100.00   &  45.74     &  100.00 \\
          & RFA   &   61.08    &   54.18    &    78.14   & 53.90      & 71.58 \\
          & Foolsgold &    71.14   &    69.77   &    99.85   &    70.52   &  99.52\\
          & MKrum &    67.87   &   66.51    &    99.89   &  64.75     &  100.00\\
          & SignGuard    &    71.22   &    70.83   &    99.26   &   70.06    & 99.78 \\
          & Deepsight &    53.36   &   55.10    &   99.35    &   54.84    &  99.99\\
          & MMetric    &   66.20    &    64.76   &   100.00    &     61.33  &  100.00\\
          & FLAME    &   65.53    &    64.84   &   99.91    &    64.16  &  100.00\\
    \bottomrule
    \end{tabular}%
    }    
  \label{tab:soda_main_table}%
\end{table}%

\textbf{Evaluated defense methods.} We evaluate the non-robust baseline \textit{FedAvg}~\cite{FL_OG} and $8$ existing SOTA defense methods, including $3$ impact reduction methods: \textit{RLR}~\cite{rlr}, \textit{RFA}~\cite{geomed}, and \textit{Foolsgold}~\cite{Foolsgold}; and $5$ filtering methods: \textit{Multi-Krum (MKrum)}~\cite{krum}, \textit{Multi-Metric (MM)}~\cite{mm}, \textit{SignGuard}~\cite{signguard}, \textit{Deepsight}~\cite{deepsight}, and \textit{FLAME}~\cite{flame}. Additionally, we compare our approach to an ideal filtering method, which perfectly identifies and removes all malicious updates, using only the benign updates to update the global model. We refer to this as the most robust baseline, \textit{FedAvg*}. Note that FedAvg* achieves the highest level of robustness theoretically. 

\textbf{Evaluation metrics.} We evaluate the performance of defense methods using: \textit{main task accuracy (MA)}, the percentage of clean test samples that are correctly classified by the global model, and \textit{attack success rate (ASR)}, the percentage of OOD test samples that are misclassified to the target label by the global model. Effective attack methods seek to maintain high MA while maximizing ASR, whereas effective defense methods aim to preserve high MA and minimize ASR, ideally yielding random guesses for OOD inputs. For filtering methods, we also measure \textit{true positive rate (TPR)}, the proportion of benign updates that are selected for aggregation in each round, and \textit{false positive rate (FPR)}, the proportion of malicious updates that are wrongly identified as benign in each round. For filtering methods, a high TPR and low FPR are preferred.
Results are shown in percentage ($\%$).


\section{Comprehensive study of \texorpdfstring{$\mathtt{SoDa}$}{SoDa}}
\textit{We first conduct comprehensive evaluation of the effectiveness of $\mathtt{SoDa}$ on attacking existing SOTA defense methods and compare $\mathtt{SoDa}$ with well-known attack counterparts in the literature. Additionally, we provide the detailed ablation study of $\mathtt{SoDa}$.}

\textbf{Effectiveness of $\mathtt{SoDa}$ on attacking existing defense methods.} 
%
We first evaluate the effectiveness of $\mathtt{SoDa}$ attacking existing SOTA defense methods. Specifically, we conduct experiments at two AR levels: $20\%$ and a more extreme $45\%$. The clean MA (without attacks), MA, and ASR of all defense methods against $\mathtt{SoDa}$ are summarized in \autoref{tab:soda_main_table}. Overall, $\mathtt{SoDa}$ consistently achieves a high ASR against all defense methods across both datasets and AR levels, demonstrating its effectiveness. Specifically, under $20\%$ AR, $\mathtt{SoDa}$ achieves nearly complete ASR against most defense methods, such as Foolsgold, MKrum, and SignGuard, on both CIFAR-10 and CIFAR-100. In a more challenging case where AR is $45\%$, the ASR reaches $100\%$ for nearly all defense methods on both datasets, indicating that a higher number of malicious clients further amplifies the attack's success. Notably, RFA shows a slightly lower ASR of $66.92\%$ at $20\%$ AR on CIFAR-10. However, as AR increases to $45\%$, it becomes ineffective against $\mathtt{SoDa}$, with ASR reaching $100\%$. $\mathtt{SoDa}$ closely aligns malicious models with benign models, thereby confusing defense methods and resulting in a high ASR. The ability of $\mathtt{SoDa}$ to maintain MA when attacking to FL system is also notable. Specifically, $\mathtt{SoDa}$ maintains an MA similar to the clean case (clean MA), indicating that $\mathtt{SoDa}$ preserves the main task accuracy while injecting the backdoor, further highlighting its effectiveness. These findings demonstrate the stealth of $\mathtt{SoDa}$.

\begin{table}[t]
  \centering
  \caption{The FPR comparison of MKrum and FLAME under $\mathtt{SoDa}$ and other attack methods on CIFAR-10 and CIFAR-100.}
  \scalebox{0.85}{
    \begin{tabular}{cc|ccccc}
    \toprule
    \textbf{Dataset} & \textbf{Method} & \textbf{Scaling} & \textbf{PGD} & \textbf{LIE} & \textbf{Neurotoxin} & $\mathtt{SoDa}$ \\
    \midrule
    \multirow{2}[1]{*}{C-10} & MKrum   & 0.00  & 5.25  & 99.25 & 44.75 & 60.50 \\
          & FLAME  & 0.00  & 8.00  & 28.00 & 68.75 & 63.25 \\
    \midrule
    \multirow{2}[1]{*}{C-100} & MKrum   & 0.00  & 79.75 & 98.25 & 50.00 & 78.75 \\
          & FLAME   & 0.00  & 89.75 & 0.00  & 68.00 & 88.50 \\
    \midrule
    \midrule
    \multicolumn{2}{c|}{\textbf{Average FPR}$\uparrow$}  & 0.00  & 45.19 & 56.88 & 57.38 & 72.25 \\
    \bottomrule
    \end{tabular}%
    }
  \label{tab:soda_attack_method_compare}%
\end{table}%

\textbf{Performance comparison between $\mathtt{SoDa}$ and SOTA attack methods.} We study the performance of some well-known backdoor attack methods when applied for $\mathtt{OBA}$ and compare them with $\mathtt{SoDa}$. 
These attack methods aim to modify local model updates to increase either the attack strength or stealthiness and can be adapted to $\mathtt{OBA}$ straightforwardly.
\autoref{tab:soda_attack_method_compare} summarizes the FPR results of $\mathtt{SoDa}$ alongside Scaling~\cite{scaling}, Projected Gradient Descent (PGD)~\cite{PGD}, Little is Enough (LIE)~\cite{lie}, and Neurotoxin~\cite{Neurotoxin} attacks, evaluated on CIFAR-10 (C-10) and CIFAR-100 (C-100) datasets. As shown, $\mathtt{SoDa}$ is more stealthy than other methods, indicated by the highest average FPR. The Scaling attack, which scales up the malicious local model updates, is easily detected by MKrum, as it relies on pair-wise Euclidean distance between local model updates. Although FLAME should theoretically remain unaffected by Scaling due to its use of Cosine similarity, in practice, the recovery process of local models incorporates the scaled updates, altering the Cosine similarity and thus making FLAME effective. Similarly, LIE, which clips malicious updates to resemble benign ones, effectively bypasses the detection of MKrum, resulting in a nearly $100\%$ FPR for both datasets. However, since LIE does not significantly alter the Cosine similarity, FLAME is able to detect it effectively. Neurotoxin is able to bypass both MKrum and FLAME, but $\mathtt{SoDa}$ achieves an average of $+14.87\%$ FPR compared to Neurotoxin, highlighting its higher attacking capabilities. 

\textbf{Ablation study of \texorpdfstring{$\mathtt{SoDa}$}{Soda}.} We study how each component functions of $\mathtt{SoDa}$. Specifically, we conduct experiments on $\mathtt{OBA}$ with single magnitude regularization term ($\mathrm{Mag}$) and single direction regularization term ($\mathrm{Dir}$). Additionally, we vary the $\lambda_m$ and $\lambda_d$ to various values. \autoref{tab: ablation_soda} summarizes the TPR and FPR of MKrum and FLAME under different attack settings on CIFAR-10 dataset. In the upper part of the table, we observe that using either the $\mathrm{Mag}$ or $\mathrm{Dir}$ component of the attack individually does not fully circumvent the defenses provided by MKrum and FLAME. However, with both $\mathrm{Mag}$ and $\mathrm{Dir}$, our $\mathtt{SoDa}$ enjoys a significant increase in FPR, reaching 61.88\%, the highest among all settings. This result implies that the combination of both components in $\mathtt{SoDa}$ effectively maximizes the attack's impact, making it the most potent strategy for overcoming the defenses. In the lower part of the table, we explore the performance of $\mathtt{SoDa}$ under varying hyperparameter settings, specifically manipulating the parameters $\lambda_m$ and $\lambda_d$. The results demonstrate that $\mathtt{SoDa}$ is highly robust to hyperparameter choices, with average TPR and FPR values remaining relatively stable across different configurations. The minimal variation in these metrics suggests that $\mathtt{SoDa}$ maintains its effectiveness regardless of specific hyperparameter choices, underscoring its adaptability.
\begin{table}[t]
  \centering
  \caption{The TPR and FPR comparison of different components of $\mathtt{SoDa}$ and $\mathtt{SoDa}$ with various coefficients.}
  \scalebox{0.85}{
    \begin{tabular}{c|cccc|cc}
    \toprule
    \multirow{2}[4]{*}{\textbf{Attack Setting}} & \multicolumn{2}{c}{\textbf{MKrum}} & \multicolumn{2}{c|}{\textbf{FLAME}} & \multirow{2}[2]{*}{\textbf{\makecell*[c]{Avg. \\ TPR$\uparrow$}}} & \multirow{2}[2]{*}{\textbf{\makecell*[c]{Avg. \\ FPR$\uparrow$}}}\\
    \cmidrule(r){2-3} \cmidrule(r){4-5} 
             & TPR$\uparrow$    & FPR$\uparrow$    & TPR$\uparrow$    & FPR$\uparrow$ \\
    \midrule
    $\mathtt{OBA}$ & 61.13  & 5.50 &   71.00  &    4.50   & 66.07 & 5.00 \\
    $\mathtt{OBA}$ w/ $\mathrm{Mag}$ & 47.06  & 61.75 &   59.50  &   20.75    & 53.28 & 41.25 \\
    $\mathtt{OBA}$ w/ $\mathrm{Dir}$  &  50.25 & 29.00 &  56.31 &    71.75   & 53.28 & 50.38 \\
    \rowcolor[rgb]{ .816,  .808,  .808} $\mathtt{SoDa}$  &  47.38 & 60.50 &  60.69 &    63.25  & 54.04 & 61.88 \\
             \midrule
    \midrule
     $\lambda_m=0.01, \lambda_d=100$ & 50.06  & 49.75 &  63.19 &   54.50    & 56.63 & 52.13 \\
      $\lambda_m=0.1, \lambda_d=10$ & 48.06  & 57.75 &  61.94  &    63.00   & 55.00 & 60.38 \\
      $\lambda_m=0.1, \lambda_d=1$ & 48.52  & 57.00 &  61.62  &   61.50    & 55.07 & 59.25 \\
      $\lambda_m=0.1, \lambda_d=0.1$ &  47.69 & 59.25 &  62.32  &    62.75   & 55.01 & 61.00 \\
    \bottomrule
    \end{tabular}%
}
  \label{tab: ablation_soda}%
\end{table}%

\section{Comprehensive Evaluation of \texorpdfstring{$\mathtt{BNGuard}$}{BNGuard}}

\textit{We evaluate the performance of $\mathtt{SoDa}$ and $\mathtt{BNGuard}$ under various settings to highlight the superior effectiveness of $\mathtt{BNGuard}$. Specifically, we assess its robustness across different data poisoning ratios and levels of data heterogeneity. We further present results on cross-device FL scenarios and conduct a component analysis to better understand the contributions of each design choice.} 

\begin{figure}
    \centering
    \includegraphics[width=0.95\linewidth]{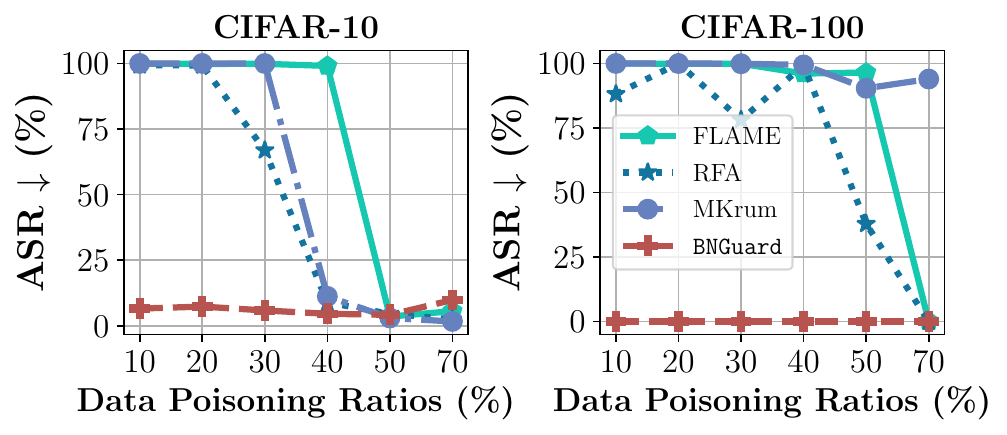}
    \caption{The ASR comparison between $\mathtt{BNGuard}$ and other methods across various data poisoning ratios under $\mathtt{SoDa}$.}
    \Description{The ASR comparison between $\mathtt{BNGuard}$ and other methods across various data poisoning ratios under $\mathtt{SoDa}$.}
    \label{fig:dpr}
\end{figure}

\textbf{Evaluation on various data poisoning ratios.} The data poisoning ratio $r$ is a critical parameter of backdoor attacks. 
A lower $r$ results in more stealthy malicious local models that are closer to benign models but may reduce the attack's effectiveness. Conversely, a higher $r$ can make malicious models more potent but causes them to deviate significantly from benign ones, making detection easier. Here, we vary the data poisoning ratio of $\mathtt{SoDa}$ from $10\%$ to $70\%$ and report the results of one impact reduction method RFA, and three filtering methods, including MKrum, FLAME, and our defense method $\mathtt{BNGuard}$, in \autoref{fig:dpr}. We observe that at a low poisoning rate $r = 10\%$, $\mathtt{SoDa}$ can achieve a \textit{complete} ASR under tested defense methods. As $r$ increases, the ASR tends to drop. In contrast to SOTA defense methods, $\mathtt{BNGuard}$ consistently achieves a low ASR across all data poisoning ratios. These results highlight not only the effectiveness of $\mathtt{SoDa}$ at low poisoning rates but also the capability of $\mathtt{BNGuard}$ in detecting stealthy malicious model updates. 

\textbf{Evaluation across different levels of data heterogeneity.} 
In practice, local clients in FL may have heterogeneous data. 
We simulate different degrees of non-IID data with $\alpha=0.7$, $0.3$, and an extreme setting of $\alpha=0.1$. \autoref{tab: non_iid} summarizes the MA and ASR results of two malicious impact reduction methods, RLR and Foolsgold, as well as three filtering methods, MKrum, FLAME, and our $\mathtt{BNGuard}$, on CIFAR-10 and CIFAR-100.
As data heterogeneity increases, local model divergence also increases, making it easier for malicious clients to hide and thus more challenging to defend against $\mathtt{SoDa}$. 
Consequently, ASR tends to rise as $\alpha$ decreases. However, in extreme non-IID settings, ASR may decrease because high data heterogeneity negatively impacts overall model performance, as indicated by a drop in MA. Overall, only $\mathtt{BNGuard}$ demonstrates robustness against $\mathtt{SoDa}$, while other methods struggle. Specifically, $\mathtt{BNGuard}$ achieves the highest backdoor robustness, with an average ASR of $7.79\%$ on CIFAR-10  and $0.04\%$ on CIFAR-100. It also maintains better MA performance; for example, on CIFAR-100, it achieves an average MA of $64.14\%$, which is $+2.07\%$ higher than the second-best method, Foolsgold. 
This demonstrates that the difference in BN layer statistics across clients, driven by OOD data, remains significant in heterogeneous data settings. In contrast, traditional metrics used in other methods become ineffective in capturing attack-induced differences, as they are overwhelmed by the variance introduced by non-IID data.

\begin{table}[t]
  \centering
  \caption{The MA and ASR comparison of SOTA methods and $\mathtt{BNGuard}$ under three data heterogeneous degrees.}
  \scalebox{0.8}{
    \begin{tabular}{c|l|cccccc|c}
    \toprule
    \multirow{2}[4]{*}{\textbf{Data.}} & \multirow{2}[4]{*}{\textbf{Method}} & \multicolumn{2}{c}{$\alpha=0.1$} & \multicolumn{2}{c}{$\alpha=0.3$} & \multicolumn{2}{c|}{$\alpha=0.7$} & \multirow{2}[2]{*}{\textbf{\makecell*[c]{Avg. \\ ASR$\downarrow$}}}\\
\cmidrule(r){3-4} \cmidrule(r){5-6} \cmidrule(r){7-8}          
& & MA$\uparrow$   & ASR$\downarrow$   & MA$\uparrow$   & ASR$\downarrow$   & MA$\uparrow$  & ASR$\downarrow$ \\
    \midrule
    \multirow{5}[2]{*}{\rotatebox{90}{CIFAR-10}} & RLR & 25.18 & 90.82 & 63.74 & 99.71 & 74.01 & 0.03 & 63.52 \\
     & Foolsgold & 20.56 & 95.76 & 85.11 & 98.74 & 43.18 & 39.74 & 78.08 \\
     & FLAME & 75.91 & 99.47 & 86.75 & 99.11 & 87.26 & 99.98 & 99.52 \\
     & MKrum & 65.32 & 84.06 & 85.13 & 99.88 & 87.75 & 97.70 & 93.88 \\
     & \cellcolor[rgb]{ .816,  .808,  .808}$\mathtt{BNGuard}$ & \cellcolor[rgb]{ .816,  .808,  .808}72.20 & \cellcolor[rgb]{ .816,  .808,  .808}8.72 & \cellcolor[rgb]{ .816,  .808,  .808}88.14 & \cellcolor[rgb]{ .816,  .808,  .808}7.90 & \cellcolor[rgb]{ .816,  .808,  .808}89.41 & \cellcolor[rgb]{ .816,  .808,  .808}6.74 & \cellcolor[rgb]{ .816,  .808,  .808}7.79 \\
    \midrule
    \midrule
    \multirow{5}[2]{*}{\rotatebox{90}{CIFAR-100}} & RLR & 23.97 & 99.93 & 29.63 & 100.00 & 38.18 & 97.24 & 99.06 \\
     & Foolsgold & 52.74 & 99.91 & 65.58 & 99.54 & 67.88 & 99.76 & 99.74 \\
     & FLAME & 50.86 & 99.99 & 61.31 & 99.94 & 63.49 & 100.00 & 99.98 \\
     & MKrum & 51.35 & 99.98 & 61.24 & 88.09 & 64.02 & 100.00 & 96.02 \\
     & \cellcolor[rgb]{ .816,  .808,  .808}$\mathtt{BNGuard}$ & \cellcolor[rgb]{ .816,  .808,  .808}58.65 & \cellcolor[rgb]{ .816,  .808,  .808}0.12 & \cellcolor[rgb]{ .816,  .808,  .808}66.07 & \cellcolor[rgb]{ .816,  .808,  .808}0.00 & \cellcolor[rgb]{ .816,  .808,  .808}67.71 & \cellcolor[rgb]{ .816,  .808,  .808}0.00 & \cellcolor[rgb]{ .816,  .808,  .808}0.04 \\
    \bottomrule
    \end{tabular}%
    }
  \label{tab: non_iid}%
\end{table}%

\textbf{Effectiveness of $\mathtt{BNGuard}$ in cross-device settings.} %
While most experiments focus on the cross-silo FL setting, evaluating the cross-device FL scenario is also essential, given the large number of clients involved.
To this end, we simulate a cross-device FL with 100 clients and the server randomly selects 20 clients to participate in training at each round. 
Here, we report the performance of FedAvg, FedAvg*, MKrum, FLAME, and $\mathtt{BNGuard}$ under $\mathtt{SoDa}$ on IID and non-IID CIFAR-10 datasets.
%
As shown in \autoref{tab: cross_device}, 
$\mathtt{SoDa}$ remains effective in cross-device settings, achieving an ASR of almost $100\%$ for FedAvg, indicating complete backdoor success. On the IID dataset, MKrum and FLAME show some resilience to $\mathtt{SoDa}$. In cross-device settings, clients have fewer samples compared to cross-silo settings, causing local models to overfit both OOD and ID samples in the poisoned local dataset. This leads to the learning of highly distinct patterns, making malicious models stand out more prominently among benign models.
However, on the non-IID dataset, $\mathtt{SoDa}$ can bypass the detection of MKrum and FLAME due to the large divergence among benign models in cross-device settings, resulting in ASRs close to $100\%$. 
In contrast, $\mathtt{BNGuard}$ achieves the lowest average ASR ($9.21\%$) and FPR ($2.02\%$), significantly outperforming other methods. It also maintains a comparable MA to FedAvg*, demonstrating strong main task performance and backdoor robustness in cross-device settings. 

\begin{table}[t]
  \centering
  \caption{Performance of different methods against $\mathtt{SoDa}$ on IID and non-IID CIFAR-10 datasets in cross-device settings.}
  \scalebox{0.88}{
    \begin{tabular}{l|cccccc|c}
    \toprule
    \multirow{2}[4]{*}{\textbf{Method}} & \multicolumn{3}{c}{\textbf{IID}} & \multicolumn{3}{c|}{\textbf{Non-IID}} & \multirow{2}[2]{*}{\textbf{\makecell*[c]{Avg. \\ FPR$\downarrow$}}}\\
    \cmidrule(r){2-4} \cmidrule(r){5-7} 
         & MA$\uparrow$    & ASR$\downarrow$    & FPR$\downarrow$    & MA$\uparrow$    & ASR$\downarrow$    & FPR$\downarrow$ \\
    \midrule
    FedAvg* &  86.13   &  10.18   &   0.00   &   81.03  &  4.57  &  0.00 & 0.00 \\
    \cmidrule{1-8}
    FedAvg &  86.26   &   99.92  &   100.00   &  81.93   &  100.00  &  100.00 & 100.00 \\
    MKrum &  85.34   &   8.30  &   1.86   &   80.14  &   99.98 &  44.47  & 23.17 \\
    FLAME &  85.98   &   10.27  &   1.23   &  79.96   & 99.99   &  64.81  & 33.02 \\
    \rowcolor[rgb]{ .816,  .808,  .808}$\mathtt{BNGuard}$  & 86.21  &  8.60 & 0.56 &  81.68   &  9.82   &    3.47       & 2.02 \\
    \bottomrule
    \end{tabular}%
}
  \label{tab: cross_device}%
\end{table}%

\textbf{Component study of $\mathtt{BNGuard}$.} The performance of $\mathtt{BNGuard}$ relies on the choice of BN statistics and clustering method. 
Here, we compare the results when using different BN layers, including the middle BN layer ($\mathtt{mid\_BN}$), the last BN layer ($\mathtt{last\_BN}$), and all BN layers combined ($\mathtt{all\_BNs}$). We also evaluate different clustering methods, including $\mathrm{DBSCAN}$~\cite{dbscan} and $\mathrm{MeanShift}$~\cite{meanshift}. In \autoref{parameter_study}, we report the TPR and FPR results on IID/non-IID CIFAR-10 and non-IID CIFAR-100 datasets. 
%
The results show that $\mathtt{BNGuard}$, which uses the statistics from the first BN layer, achieves the highest average TPR of $99.23\%$ and a low average FPR of $1.50\%$. In contrast, using statistics from other BN layers, such as $\mathtt{mid\_BN}$ and $\mathtt{last\_BN}$, results in lower TPR values (e.g., $66.85\%$ and $73.46\%$) and higher FPR values (e.g., 51.58\% and 54.92\%), indicating that these statistics are less effective at capturing the patterns of OOD data. Using statistics from all BN layers ($\mathtt{all\_BNs}$) achieves a performance level similar to that of the last BN layer, though it is less effective compared to using only the first BN layer. These results suggest that the statistics from the first BN layer are the most effective for accurately identifying benign clients while minimizing the selection of malicious ones, making it a practical choice for deploying $\mathtt{BNGuard}$.
%
The results in \autoref{parameter_study} also demonstrate that $\mathtt{BNGuard}$ enjoys a stable performance across different clustering methods. Specifically, with either $\mathrm{DBSCAN}$ or $\mathrm{MeanShift}$, the average TPR remains high, and the average FPR stays low, indicating that the choice of clustering method does not significantly affect the effectiveness of $\mathtt{BNGuard}$. 
This stability is due to the clear distinction in BN statistics between malicious and benign models. 
Overall, these results highlight that using statistics from the first BN layer enables $\mathtt{BNGuard}$ to identify the malicious models in $\mathtt{SoDa}$ effectively. 
%

\begin{table}[t]
  \centering
  \caption{The TPR and FPR of $\mathtt{BNGuard}$ with different components on CIFAR-10 and CIFAR-100 datasets.}
  \label{parameter_study}
  \scalebox{0.8}{
    \begin{tabular}{l|cccccc|cc}
    \toprule
    \multirow{2}[4]{*}{\textbf{Setting}} & \multicolumn{2}{c}{\textbf{C-10}} & \multicolumn{2}{c}{\textbf{$\text{C-10 (N)}$}} & \multicolumn{2}{c|}{\textbf{$\text{C-100 (N)}$}} & \multirow{2}[2]{*}{\textbf{\makecell*[c]{Avg. \\ TPR$\uparrow$}}} & \multirow{2}[2]{*}{\textbf{\makecell*[c]{Avg. \\ FPR$\downarrow$}}}\\
\cmidrule(r){2-3} \cmidrule(r){4-5} \cmidrule(r){6-7}          & TPR$\uparrow$   & FPR$\downarrow$   & TPR$\uparrow$   & FPR$\downarrow$   & TPR$\uparrow$  & FPR$\downarrow$ \\
    \midrule
    $\mathtt{mid\_BN}$ & 71.19 & 41.25 & 65.81 & 49.25 & 63.56  & 64.25 & 66.85 & 51.58 \\
    $\mathtt{last\_BN}$ & 60.56 & 79.75 & 80.87 & 74.25  & 78.94 & 10.75 & 73.46 & 54.92 \\
    $\mathtt{all\_BNs}$ & 97.94 & 3.75 & 59.38 & 85.00 & 60.75  & 71.00 & 72.69 & 53.25 \\
    \midrule
    \midrule
    $\mathrm{DBSCAN}$ & 99.69 & 0.00 & 94.00 & 0.00  & 99.69 & 36.00 & 97.79 & 12.00 \\
    $\mathrm{MeanShift}$ & 100.00 & 0.75 & 97.81 & 0.50  & 98.00 & 1.50 & 98.60 & 0.92 \\
    \midrule
    \rowcolor[rgb]{ .816,  .808,  .808}$\mathtt{BNGuard}$  & 100.00 &  0.75 & 98.88 & 1.25  & 98.81 & 2.50 & 99.23 & 1.50 \\
    \bottomrule
    \end{tabular}%
    }
\end{table}%


\section{More Discussions}

\begin{table}[t]
  \centering
  \caption{The performance of MKrum, FLAME, and $\mathtt{BNGuard}$ with FMNIST and SVHN as the OOD dataset for $\mathtt{SoDa}$.}
  \scalebox{0.8}{
    \begin{tabular}{c|l|cccccc|c}
    \toprule
    \multirow{2}[2]{*}{\textbf{\makecell*[c]{ID \\ Dataset}}} & \multirow{2}[4]{*}{\textbf{Method}} & \multicolumn{3}{c}{\textbf{FMNIST}} & \multicolumn{3}{c|}{\textbf{SVHN}} & \multirow{2}[2]{*}{\textbf{\makecell*[c]{Avg. \\ FPR$\downarrow$}}}\\
    \cmidrule(r){3-5} \cmidrule(r){6-8} 
         & & MA$\uparrow$    & ASR$\downarrow$    & FPR$\downarrow$    & MA$\uparrow$    & ASR$\downarrow$    & FPR$\downarrow$ \\
    \midrule
    \multirow{3}[2]{*}{\rotatebox{90}{GTSRB}} & MKrum &   96.05  &  12.38  &  20.50  & 95.53  & 93.91   &  89.50 & 55.03 \\
     & FLAME &   95.85  &  24.53  &  34.75  & 97.06  & 96.48   &  94.50 & 64.70 \\
     & \cellcolor[rgb]{ .816,  .808,  .808}$\mathtt{BNGuard}$  & \cellcolor[rgb]{ .816,  .808,  .808}96.36  & \cellcolor[rgb]{ .816,  .808,  .808}2.33 & \cellcolor[rgb]{ .816,  .808,  .808}0.00 &  \cellcolor[rgb]{ .816,  .808,  .808}97.55  &  \cellcolor[rgb]{ .816,  .808,  .808}5.11  &    \cellcolor[rgb]{ .816,  .808,  .808}1.50     & \cellcolor[rgb]{ .816,  .808,  .808}0.75 \\
        \midrule
    \midrule
    \multirow{3}[2]{*}{\rotatebox{90}{C-10}} & MKrum &  90.58   &   99.22 &   60.50   &  91.15  &  98.32  &  67.25  & 63.88 \\
     & FLAME & 90.88  & 98.61 & 64.75 &  91.26  &  93.20  &    66.50     & 65.63 \\
     & \cellcolor[rgb]{ .816,  .808,  .808}$\mathtt{BNGuard}$  & \cellcolor[rgb]{ .816,  .808,  .808}91.09  & \cellcolor[rgb]{ .816,  .808,  .808}9.91 & \cellcolor[rgb]{ .816,  .808,  .808}1.00 &  \cellcolor[rgb]{ .816,  .808,  .808}91.14  &  \cellcolor[rgb]{ .816,  .808,  .808}0.14  &    \cellcolor[rgb]{ .816,  .808,  .808}1.25     & \cellcolor[rgb]{ .816,  .808,  .808}1.13 \\
        \midrule
    \midrule
    \multirow{3}[2]{*}{\rotatebox{90}{C-100}} & MKrum &  66.33   &  99.58  &    78.25  &  66.24  &   98.87 &  80.00  & 79.13 \\
     & FLAME & 65.07  & 99.39 &  89.00 &  64.70  &  98.19  &     91.00    & 90.00 \\
     & \cellcolor[rgb]{ .816,  .808,  .808}$\mathtt{BNGuard}$  &  \cellcolor[rgb]{ .816,  .808,  .808}69.61 & \cellcolor[rgb]{ .816,  .808,  .808}0.00 & \cellcolor[rgb]{ .816,  .808,  .808}3.50 &  \cellcolor[rgb]{ .816,  .808,  .808}69.69  &  \cellcolor[rgb]{ .816,  .808,  .808}0.02  &    \cellcolor[rgb]{ .816,  .808,  .808}3.25     & \cellcolor[rgb]{ .816,  .808,  .808}3.38 \\
    \bottomrule
    \end{tabular}%
}
  \label{tab: different_ood}%
\end{table}%

\textbf{Evaluation on \texorpdfstring{$\mathtt{SoDa}$}{SoDa} with different OOD datasets.} \label{apdx: soda_other_ood_dataset}
The performance of various methods under $\mathtt{SoDa}$ using FMNIST and SVHN as OOD datasets is shown in \autoref{tab: different_ood}. The results demonstrate that $\mathtt{SoDa}$ remains effective in maintaining high ASR when using diverse OOD datasets, as seen in the results of MKrum and FLAME. Specifically, both MKrum and FLAME exhibit high ASR values close to $100\%$ across FMNIST and SVHN on both CIFAR-10 and CIFAR-100 datasets, indicating that the backdoor remains successful even when different OOD data are used. However, $\mathtt{BNGuard}$ shows consistent robustness against these various datasets. It achieves significantly lower ASR values compared to MKrum and FLAME, with ASR values of $9.91\%$ and $0.14\%$ on CIFAR-10, and $0.00\%$ and $0.02\%$ on CIFAR-100 for FMNIST and SVHN, respectively. This is because $\mathtt{BNGuard}$ maintains low FPRs across different OOD settings, with average FPRs of $1.13\%$ on CIFAR-10, and $3.38\%$ on CIFAR-100, highlighting its ability to effectively differentiate between malicious and benign updates without overly penalizing benign clients. For the more practical dataset GTSRB, when using FMNIST as the OOD dataset, both MKrum and FLAME exhibit a certain degree of robustness. However, $\mathtt{BNGuard}$ still outperforms them. When SVHN is used as the OOD dataset, $\mathtt{BNGuard}$ achieves the best performance, whereas MKrum and FLAME completely lose their robustness. These results indicate that while $\mathtt{SoDa}$ remains a challenging backdoor attack under various OOD conditions, $\mathtt{BNGuard}$ effectively mitigates such attacks, providing a robust defense mechanism across different datasets and OOD triggers.

\textbf{Alternatives of self-reference training in \texorpdfstring{$\mathtt{SoDa}$}{SoDa}.} \label{apdx: limitation_and_future_work_of_soda}
As intuitive, $\mathtt{SoDa}$ leverages a self-reference training phase to calculate the benign reference model, which needs an extra training time for malicious clients compared with benign ones. This extra training time may limit the effectiveness of $\mathtt{SoDa}$ in \textit{synchronized FL settings}, as clients unable to return their local model updates on time are disregarded. Here, we discuss several alternatives to obtain a reference local model instead of using the self-reference training phase. 

One potential solution is for malicious clients to use the downloaded global model as a benign reference. However, if the global model has already been compromised, this approach may render regularization ineffective, as the aim of regularization is to statistically align the malicious local model with a benign one. We denote this method as $\mathtt{SoDa\_A}$. Another possible solution is for malicious clients to start training a reference benign model with the current round’s downloaded global model immediately after uploading their local model updates. When the next round's global model arrives, they can use this pre-trained reference model, utilizing gap time effectively. We denote this method as $\mathtt{SoDa\_B}$. We summarize the MA, ASR, and FPR for MKrum, FLAME, and $\mathtt{BNGuard}$ under $\mathtt{SoDa\_A}$ and $\mathtt{SoDa\_B}$ in \autoref{tab: different_soda}. 
Our results indicate that $\mathtt{SoDa\_B}$ effectively attacks both MKrum and FLAME, achieving maximum ASR. This suggests that $\mathtt{SoDa\_B}$ can serve as an alternative to $\mathtt{SoDa}$, resulting in an effective attack strategy in synchronized FL settings. In contrast, $\mathtt{SoDa\_A}$ is effective primarily on the CIFAR-100 dataset but shows limited impact on CIFAR-10. Notably, $\mathtt{BNGuard}$ demonstrates robust defense across both attack scenarios. We encourage further exploration of strategies to enhance $\mathtt{SoDa}$.

\begin{table}[t]
  \centering
  \caption{The MA, ASR, and FPR comparison of MKrum, FLAME, and $\mathtt{BNGuard}$ under $\mathtt{SoDa\_A}$ and $\mathtt{SoDa\_B}$.}
  \scalebox{0.8}{
    \begin{tabular}{c|l|cccccc|c}
    \toprule
    \multirow{2}[4]{*}{\textbf{Data}} & \multirow{2}[4]{*}{\textbf{Method}} & \multicolumn{3}{c}{\textbf{$\mathtt{SoDa\_A}$}} & \multicolumn{3}{c|}{\textbf{$\mathtt{SoDa\_B}$}} & \multirow{2}[2]{*}{\textbf{\makecell*[c]{Avg. \\ FPR$\downarrow$}}}\\
    \cmidrule(r){3-5} \cmidrule(r){6-8} 
         & & MA$\uparrow$    & ASR$\downarrow$    & FPR$\downarrow$    & MA$\uparrow$    & ASR$\downarrow$    & FPR$\downarrow$ \\
    \midrule
    \multirow{3}[2]{*}{\rotatebox{90}{C-10}} & MKrum & 90.63 & 3.63 & 4.50  & 90.56 & 99.99  &  54.75 &  29.63 \\
     & FLAME & 90.21 & 8.19 &  9.25 & 90.51 & 99.77  & 57.25  &  33.25 \\
     & \cellcolor[rgb]{ .816,  .808,  .808}$\mathtt{BNGuard}$  & \cellcolor[rgb]{ .816,  .808,  .808}91.35 & \cellcolor[rgb]{ .816,  .808,  .808}8.11 & \cellcolor[rgb]{ .816,  .808,  .808}0.75  & \cellcolor[rgb]{ .816,  .808,  .808}91.23 &  \cellcolor[rgb]{ .816,  .808,  .808}4.27 &  \cellcolor[rgb]{ .816,  .808,  .808}0.75 &  \cellcolor[rgb]{ .816,  .808,  .808}0.75 \\
        \midrule
    \midrule
    \multirow{3}[2]{*}{\rotatebox{90}{C-100}} & MKrum & 65.31 & 99.99 & 78.50  & 66.77 &  99.99 & 72.00  &  75.25 \\
     & FLAME & 64.67 & 99.93 & 87.50  & 64.97 & 99.93  &  88.00 &  87.75 \\
     & \cellcolor[rgb]{ .816,  .808,  .808}$\mathtt{BNGuard}$  & \cellcolor[rgb]{ .816,  .808,  .808}69.37 &  \cellcolor[rgb]{ .816,  .808,  .808}0.00&  \cellcolor[rgb]{ .816,  .808,  .808}3.50 & \cellcolor[rgb]{ .816,  .808,  .808}69.67 &  \cellcolor[rgb]{ .816,  .808,  .808}0.00 & \cellcolor[rgb]{ .816,  .808,  .808}3.75  &  \cellcolor[rgb]{ .816,  .808,  .808}3.63 \\
    \bottomrule
    \end{tabular}%
}
  \label{tab: different_soda}%
\end{table}%

\textbf{Effectiveness of \texorpdfstring{$\mathtt{BNGuard}$}{BNGuard} on different models.} \label{apdx: bnguard_on_various_models}
We use three benchmark deep learning models to comprehensively demonstrate the robust defense performance of $\mathtt{BNGuard}$. Specifically, we use DenseNet-121~\cite{densenet121}, MobileNetv2~\cite{mobilenetv2}, and VGG16~\cite{vgg16_bn} on CIFAR-10 and CIFAR-100 datasets, and summarize ASR and FPR of $\mathtt{BNGuard}$ in Table~\ref{tab: model_study}. The results show that $\mathtt{BNGuard}$ maintains low ASR and FPR values across various models and datasets, indicating its robust performance in mitigating backdoor attacks. For instance, $\mathtt{BNGuard}$ consistently achieves low ASR values, with an average ASR of $5.95\%$ on CIFAR-10 and $0.00\%$ on CIFAR-100, demonstrating its ability to defend $\mathtt{SoDa}$. Similarly, the FPR values remain controlled, with an average FPR of $3.08\%$ on CIFAR-10 and $2.17\%$ on CIFAR-100, ensuring a minimal false positive detection rate. These results indicate that $\mathtt{BNGuard}$ provides a reliable defense across different architectures and datasets, making it a versatile solution against backdoor attacks in FL.

\begin{table}[t]
  \centering
  \caption{The ASR and FPR of $\mathtt{BNGuard}$ with different models on CIFAR-10 and CIFAR-100 datasets.}
  \label{tab: model_study}
  \scalebox{0.8}{
    \begin{tabular}{c|cccccc|cc}
    \toprule
    \multirow{2}[4]{*}{\textbf{Data}} & \multicolumn{2}{c}{\textbf{DenseNet121}} & \multicolumn{2}{c}{\textbf{MobileNetv2}} & \multicolumn{2}{c|}{\textbf{VGG16}} & \multirow{2}[2]{*}{\textbf{\makecell*[c]{Avg. \\ ASR$\downarrow$}}} & \multirow{2}[2]{*}{\textbf{\makecell*[c]{Avg. \\ FPR$\downarrow$}}}\\
\cmidrule(r){2-3} \cmidrule(r){4-5} \cmidrule(r){6-7}          & ASR$\downarrow$   & FPR$\downarrow$   & ASR$\downarrow$   & FPR$\downarrow$   & ASR$\downarrow$  & FPR$\downarrow$ \\
    \midrule
    C-10 & 2.02 & 0.50 & 13.31 & 8.75 & 2.52  & 0.00 & 5.95 & 3.08 \\
        \midrule
    \midrule
    C-100 & 0.00 & 0.25 & 0.00 & 7.25 &  0.00 & 1.00 & 0.00 & 2.17 \\
    \bottomrule
    \end{tabular}%
    }
\end{table}%

\textbf{Execute time of different defense methods.}
\label{apdx: computational_cost}
As each step in $\mathtt{BNGuard}$ requires only a constant runtime, it is a time-efficient defense method that enhances its practicality. We summarized the average execution times for MKrum, FLAME, Foolsgold, Deepsight, SignGuard, and $\mathtt{BNGuard}$ in \autoref{tab: ex_time}. The results indicate that $\mathtt{BNGuard}$'s execution time is comparable to other SOTA methods, completing detection in just 0.3 seconds. This time efficiency makes $\mathtt{BNGuard}$ a practical choice for deployment.

\begin{table}[t]
  \centering
  \caption{Execution time (in seconds) of different methods on CIFAR-10 dataset.}
  \scalebox{0.8}{
    \begin{tabular}{cccccc}
    \toprule
     \textbf{FedAvg} & \textbf{MKrum} & \textbf{FLAME} & \textbf{SignGuard} & \textbf{Deepsight} &  $\mathtt{BNGuard}$ \\
    \midrule
     0.0048 & 0.12 & 0.30 & 0.47 & 365.81 &  0.30 \\
    \bottomrule
    \end{tabular}%
  }
  \label{tab: ex_time}%
\end{table}

\section{Conclusion}
In this work, we introduce a novel backdoor attack prototype, $\mathtt{OBA}$, which significantly enhances the practicality of backdoor attacks in FL. Unlike traditional ID backdoor attacks that rely on unnatural and visually conspicuous triggers embedded in data samples, $\mathtt{OBA}$ enables arbitrary objects to act as potential triggers. This substantially broadens the range of attack scenarios while improving stealth and flexibility. To further enhance the stealthiness of $\mathtt{OBA}$, we propose $\mathtt{SoDa}$, a method that aligns the malicious local model with its benign counterpart, thereby evading existing SOTA defense mechanisms. We conduct comprehensive experiments demonstrating the effectiveness of $\mathtt{SoDa}$ in bypassing 8 SOTA defenses, as indicated by its nearly perfect ASR while maintaining MA. To address this emerging and critical security threat, we propose a novel defense method, $\mathtt{BNGuard}$, which detects malicious behavior by examining the running statistics from the first BN layer of local models. $\mathtt{BNGuard}$ is particularly targeted at defending FL systems with models that contain BN layers. Extensive experiments validate the effectiveness and robustness of $\mathtt{BNGuard}$, showing that it consistently outperforms existing defense methods and robustly mitigates the impact of $\mathtt{SoDa}$ across various scenarios.

\begin{acks}
This work was supported by the National Science Foundation RII Track-1: Harnessing the Data Revolution for Fire Science (HDRFS) Seed Grants under NSHE-24-37.
\end{acks}

\printbibliography

\end{document}